\documentclass[11pt]{article}
\usepackage{coling2020}
\usepackage{times}
\usepackage{url}
\usepackage[utf8]{inputenc}
\usepackage{latexsym}
\usepackage{multirow}
\usepackage{graphicx}
\usepackage[textsize=scriptsize,backgroundcolor=green!20]{todonotes}
\usepackage{subfig}

\usepackage{microtype}

\newcommand{\revisar}[1]{{\color{black}#1}}
\newcommand{\revisarcr}[1]{#1}

\colingfinalcopy 

\title{Understanding the effects of word-level linguistic annotations in under-resourced neural machine translation}

\author{
Víctor M. Sánchez-Cartagena,
Juan Antonio Pérez-Ortiz,
Felipe Sánchez-Martínez\\[1ex]
Dep. de Llenguatges i Sistemes Informàtics,
Universitat d'Alacant\\
E-03690 Sant Vicent del Raspeig (Spain)\\
{\tt \{vmsanchez,japerez,fsanchez\}@dlsi.ua.es}}

\date{}

\begin{document}
\maketitle
\begin{abstract}
This paper studies the effects of word-level linguistic annotations in under-resourced neural machine translation, for which \revisar{there is incomplete evidence in the literature}.
The study covers eight language pairs, different training corpus sizes, two architectures and three types of annotation: dummy tags (with no linguistic information at all), part-of-speech tags, and morpho-syntactic description tags, which consist of part of speech and morphological features. These linguistic annotations are \emph{interleaved} in the input or output streams as a single tag placed before each word. In order to measure the performance under each scenario, we use automatic evaluation metrics and perform automatic error classification. Our experiments show that, in general, source-language annotations are helpful and morpho-syntactic descriptions outperform part of speech for some language pairs. On the contrary, when words are annotated in the target language, part-of-speech tags systematically outperform morpho-syntactic description tags in terms of automatic evaluation metrics, even though the use of \revisar{morpho-syntactic description tags} improves the grammaticality of the output. We provide a detailed analysis of the reasons behind this result.

\end{abstract}

\section{Introduction}

\blfootnote{
    %
    %
    %
    
     \hspace{-0.65cm}  
     This work is licensed under a Creative Commons 
     Attribution 4.0 International Licence.
     Licence details:
     \url{http://creativecommons.org/licenses/by/4.0/}.
    %
    %
}

Training neural machine translation (NMT) systems for under-resourced language pairs, for which the amount of parallel corpora is orders of magnitude smaller than those available for prevailing language pairs, may be challenging.  Recently, \newcite{sennrich-zhang-2019-revisiting} have shown that even under these circumstances, NMT surpasses classical approaches such as phrase-based statistical MT \cite{koehnbook}. In these under-resourced scenarios, the use of 
relevant linguistic word-level annotations 
has proved to improve translation  performance~\cite{sennrich-haddow-2016-linguistic,nadejde:2017}.

Linguistic annotations can be used to label source-language (SL) or target-language (TL) words. The former lead to more accurate representations of the SL sentence~\cite{sennrich-haddow-2016-linguistic} and usually require changes in the encoder but not in the training loss~\cite{chen-etal-2017-neural,li-knowledge-aware}.
The latter involve producing probability distributions for both TL words and TL linguistic annotations, which can be seen as a form of multi-task learning. Multi-task learning architectures explored in the literature include: independent decoders for words and linguistic annotations~\cite{zhou-etal-2017-chunk,10.1109/TASLP.2018.2855968,gu-etal-2018-top,wang-etal-2018-tree,yang-etal-2019-latent}; independent output layers in the same decoder~\cite{garcia-martinez-barrault-bougare:2016,gronroos-etal-2017-extending,feng-improved-nmt}; and even sharing the same network for both tasks~\cite{nadejde:2017,tamchyna-etal-2017-modeling,wagner2017} and alternatively produce linguistic annotations and words. The latter approach is usually referred to as \emph{interleaving}.

We can also classify the approaches according to the type of linguistic annotations used: part-of-speech tags~\cite{feng-improved-nmt,yang-etal-2019-latent}; morpho-syntactic description tags, which comprise part of speech and morphological inflection information~\cite{garcia-martinez-barrault-bougare:2016,tamchyna-etal-2017-modeling}; and syntactic structure information~\cite{nadejde:2017,chen-etal-2017-neural,zhou-etal-2017-chunk,10.1109/TASLP.2018.2855968,gu-etal-2018-top,wang-etal-2018-tree}. Using morpho-syntactic description tags as TL annotations allows us to train the network to produce TL lemmas instead of surface forms (words as they appear in running texts). This strategy can reduce data sparseness but requires the use of an external morphological generator as a post-processing step~\cite{tamchyna-etal-2017-modeling}.

Despite the body of work published about this topic, no strategy has clearly emerged as the most appropriate method for integrating linguistic annotations into  NMT. The literature mainly contains incomplete evidence. For instance, \newcite{yang-etal-2019-latent} conclude that TL part-of-speech annotations boost translation quality with an \emph{ad-hoc} architecture, but \newcite{wagner2017} claims that TL morpho-syntactic description tags degrade translation quality when they are interleaved: it is not clear whether the difference between both results is caused by the type of linguistic annotations or by the approach followed to integrate them. There are also contradictory results, such as those reported by \newcite{tamchyna-etal-2017-modeling}, who claim that TL annotations are only useful when they are combined with lemmatisation, and \newcite{nadejde:2017}, who report positive results without lemmatisation. In addition, the influence of factors such as the size of the available training parallel corpus and the language typology have not been properly evaluated.


\revisar{In this paper, we aim at clarifying how linguistic annotations help NMT by carrying out systematic experiments with eight language pairs. 
We focus on an under-resourced scenario where linguistic annotations are likely to provide information that cannot be inferred from scarce training data.}
We analyse multiple factors, namely, language typology, side which is annotated with linguistic information (SL, TL or both), architecture of the NMT system, training corpus size, and type of information encoded in the tags. For the latter factor, we focus only on part-of-speech tags and morpho-syntactic description tags, since other type of annotations, such as 
CCG supertags~\cite{steedman2000syntactic}, are unlikely to be available for under-resourced languages. 
\revisar{We train systems on linguistically annotated surface forms via interleaving, which does not require modifications to the neural network and allows us to easily compare different types of linguistic annotations and NMT architectures. }
In addition, an automatic error classification allows us to qualitatively compare them. A qualitative analysis has only been previously performed by \newcite{nadejde:2017}, but it covered only two language pairs with English as TL, and a single type of \revisar{annotation}.

The rest of the paper is organised as follows. Next section overviews the process of interleaving linguistic annotations in the 
training data. Section \ref{se:experiments} then describes the experimental settings 
whereas Section \ref{se:results} reports and discusses the results obtained. Section~\ref{sec:analysis} presents automatic error classification results 
for all the systems evaluated, while Section~\ref{sec:tlmorph} studies the reasons behind the poor performance of systems with TL morpho-syntactic description tags. The paper ends with some concluding remarks. 

\section{Interleaving in neural machine translation} \label{se:interleaving}
 
The interleaving approach for integrating linguistic annotations into NMT~\cite{nadejde:2017} annotates each word with a single tag which is \emph{interleaved} in the sentence before the word, i.e. introduced in the sentence as if it were another word. In our experiments, tags can represent either the part of speech (POS) of the word or its morpho-syntactic description (MSD). 
As corpora are pre-processed with BPE~\cite{sennrich-haddow-birch:2016}, the tag is introduced just once, before the first sub-word unit. To study if the effect of using tags is related to the fact that input and output sequences get longer and word boundaries are explicitly defined, and not to the information provided by tags, we also tried with a dummy tag (DUM) conveying no linguistic information at all, and used the same dummy tag for every word~\cite{wagner2017}. \revisarcr{Interleaved TL tags are removed from the final translation generated by the system before computing the automatic evaluation metrics.}

The example below shows the result of interleaving MSD tags in the English sentence \emph{It has happened before}. The sentence contains a pronoun (\texttt{PRON}) followed by an auxiliary verb (\texttt{AUX}), a main verb (\texttt{VERB}), an adverb (\texttt{ADV}) and a punctuation mark (\texttt{PUNCT}). The analysis of the pronoun tells us that it is personal, nominative, neuter, singular, and 3rd-person. The symbol \texttt{@@} acts as a sub-word unit separator.

\vspace{0.1cm}
\texttt{PRON\_Case=Nom|Gender=Neut|Number=Sing|Person=3|PronType=Prs} it

\texttt{AUX\_Mood=Ind|Number=Sing|Person=3|Tense=Pres|VerbForm=Fin} has

\texttt{VERB\_Tense=Past|VerbForm=Part} happen$@@$ ed \texttt{ADV\_\_} before \texttt{PUNCT\_\_} .

\section{Experimental settings}\label{se:experiments}

We conducted experiments for the translation of English text into four languages, and vice-versa. These languages ---Czech (\texttt{cs}), German (\texttt{de}), Spanish (\texttt{es}) and Turkish (\texttt{tr})--- belong to different language families and differ at the syntactic and morphological levels. German, Czech and Spanish are Indo-European languages: they are, respectively, Germanic, Slavic and Romance. \revisar{Both German and Czech have declension and SVO sentence structure, except for the subordinate sentences in German, which are SOV. 
Spanish has no declension and its sentence structure is SVO. Turkish is an agglutinative Turkic language with declension and SOV sentence structure.}  The morphological differences between these languages are reflected in the sparsity of the MSD tags: the number of unique tags in the interleaved training corpora ranges from a few hundreds for English and Spanish to a few thousands for Czech, German and Turkish.\footnote{The exact figures are as follows: English: 208; Czech: 1,904; German: 1,077; Spanish: 365; Turkish: 1,859.}


We simulated an under-resourced scenario by downsampling available parallel corpora for the selected language pairs. Downsampling has important advantages over using truly under-resourced language pairs: (i) we can choose languages from different families and evaluate them on standard, high-quality test sets; 
(ii) we can 
confirm whether conclusions hold for richer-resource scenarios by training on larger datasets for the same language pairs; and (iii) linguistic annotations can be obtained with the same state-of-the-art morphological analyser, minimising the potential distortions introduced by differences in the morphological analyser technology and in performance between the languages. The POS and MSD tags were obtained by means of the StandfordNLP tagger~\cite{qi2018universal}. 
In any case, the approaches described in this paper could be applied to truly under-resourced language pairs as transfer learning allows to obtain morphological analysers even from scarce morphologically annotated data~\cite{kondratyuk-2019-cross}.


\paragraph{Corpora.}

The training, development and test sets used belong to the news domain. For training, we used texts from the News Commentary v14 corpus,\footnote{\url{http://data.statmt.org/news-commentary/v14/}} except for Turkish, for which we used texts from the SETimes corpus~\cite{tyers2010south}. For development and testing we used evaluation sets from the WMT 2019 Conference on Machine Translation, \revisar{each of which contains around 3,000 parallel sentences}.\footnote{For Turkish, German and Czech we used newstest2017 and newstest2018 for development and testing, respectively. For Spanish, we used newstest2012 for development and newstest2013 for testing. } 
To see if the conclusions drawn on the under-resourced settings hold in a richer-resourced scenario, we trained English--German systems (in both directions) on the concatenation of the parallel data made available for the WMT 2017 shared task on news translation\footnote{\url{http://data.statmt.org/wmt17/translation-task/preprocessed}} plus the 
synthetic parallel data \revisar{obtained through back-translation} 
released
by~\newcite{sennrich-etal-2016-edinburgh}.

All corpora were tokenised and truecased with the Moses scripts\footnote{\url{https://github.com/moses-smt/mosesdecoder/tree/master/scripts}} and parallel sentences longer than 100 words in either side were discarded. Table \ref{tb:corpora} provides information about 
the training corpora after their pre-processing. We trained translation models on these corpora and on random sub-sets of them containing 50k parallel sentences (except for the WMT training data). The token counts depicted in Table~\ref{tb:corpora} for the under-resourced scenario are similar to those listed in the OPUS collection~\cite{TIEDEMANN12.463} for under-resourced language pairs such as English--Kurdish or English--Igbo; token counts for the 50k subsets match other pairs with even smaller resources available in OPUS, such as English--Kazakh.

\begin{table}
\begin{small}
\centering
\revisarcr{
\begin{tabular}{lcc|lcc}
\textbf{language pair} & \textbf{tokens} & \textbf{sentences} & \textbf{language pair} & \textbf{tokens} & \textbf{sentences}\\
\hline
\multicolumn{3}{c}{\bf under-resourced} & \multicolumn{3}{c}{\bf WMT}  \\
\hline
English--Czech & 6.0M -- \phantom{1}5.4M & 240k & English--German & 212M -- 200M & 9.4M\\
English--German & 8.4M -- \phantom{1}8.4M & 329k& German--English & 232M -- 240M & 10.0M \\
English--Spanish & 9.5M -- 10.6M & 367k & & &\\
English--Turkish & 5.2M -- \phantom{1}4.7M & 208k & & &\\
\hline
\end{tabular}
}
\caption{\label{tb:corpora} Number of sentences and tokens in the parallel corpora used for training. \revisarcr{The column labelled as \emph{tokens} depicts the number of tokens for each language of the pair, in the order defined by the content of the  column \emph{language pair}}. Figures for the WMT data differ between both directions because of the use of backtranslated corpora. }
\end{small}
\end{table}

\paragraph{Translation models.}
We tested the performance of the recurrent-neural-network encoder-decoder with attention (hereafter,  \emph{recurrent}; Bahdanau et al., 2015) and the Transformer \cite{NIPS2017_7181} architectures 
when the different types of tags introduced in Section \ref{se:interleaving} are interleaved in the SL input sequence, in the TL output sequence, and in both of them. For each architecture, we also trained a baseline using no tags at all. To keep the experiments to a manageable size, the systems that included tags in both languages were not trained on the nine possible combinations of tag types (three in the SL and three in the TL).
Instead, they were trained only on SL MSD and TL POS tags, which were those with the best general performance when used in isolation in the SL and in the TL, respectively. For the same reason, we only explored the SL MSD/TL POS tag combination for systems trained on large-scale WMT data. 






In order to determine the appropriate values for training hyper-parameters, a grid search over the number of BPE operations and the neural network sizes was carried out. The optimum hyper-parameter values for each language pair, training corpus size and architecture were obtained after training the baseline systems. These hyper-parameters were also used with the systems integrating linguistic annotations. Appendix \ref{se:hyperparam} provides 
a detailed description of the training process.




\paragraph{Error classification.}
We followed the automatic error analysis strategy by \newcite{toral-sanchez-cartagena-2017-multifaceted}, 
who used the tool Hjerson~\cite{popovic2011hjerson} to classify 
word errors into five categories: inflection, 
reordering, 
missing words, extra words and incorrect lexical choices. As it is difficult to automatically distinguish between the latter three categories \cite{Popovic:2011:TAE:2077692.2077694}, we grouped them into a unique category named \emph{lexical errors}. Hjerson works on the surface form and lemma of the words in the reference translations and MT outputs. The lemmas were obtained again with the StandfordNLP tagger.

\section{Results and discussion}\label{se:results}
\label{sec:results}

\renewcommand{\arraystretch}{1.0}
\begin{table*}[ht!]
\centering
\footnotesize \addtolength{\tabcolsep}{-1.7pt}
    \begin{tabular}{c|l||r|r||r|r||r|r||r|r}
    \multirow{3}{*}{\textbf{Language}} &  \multirow{3}{*}{\textbf{Tags}} & \multicolumn{4}{c||}{\bf English as SL} & \multicolumn{4}{c}{\bf English as TL}  \\
    \cline{3-10}
    & & \multicolumn{2}{c||}{\bf recurrent} & \multicolumn{2}{c||}{\bf Transformer} & \multicolumn{2}{c||}{\bf recurrent} & \multicolumn{2}{c}{\bf Transformer} \\
    & & \multicolumn{1}{c|}{\textbf{50k}}  &  \multicolumn{1}{c||}{\textbf{Full}}   &  \multicolumn{1}{c|}{\textbf{50k}}  &  \multicolumn{1}{c||}{\textbf{Full}} &  \multicolumn{1}{c|}{\textbf{50k}}  &  \multicolumn{1}{c||}{\textbf{Full}}   &  \multicolumn{1}{c|}{\textbf{50k}}  &  \multicolumn{1}{c}{\textbf{Full}}  \\
    \hline
     \multirow{8}{*}{Czech} &  None (baseline) & 7.59\phantom{$\bullet\dagger$}  &  11.98\phantom{$\bullet\dagger$} &  6.97\phantom{$\bullet\dagger$}  &  12.35\phantom{$\bullet\dagger$}   & 12.34\phantom{$\bullet\dagger$}  & 17.81\phantom{$\bullet\dagger$}   &  10.63\phantom{$\bullet\dagger$}  &   17.58\phantom{$\bullet\dagger$}  \\
     &  SL DUM & 7.61\phantom{$\bullet\dagger$}   &   12.30\phantom{$\bullet\dagger$} & 6.87\phantom{$\bullet\dagger$} & 12.66\phantom{$\bullet\dagger$} &  \textbf{12.79}\phantom{$\bullet\dagger$} & 17.98\phantom{$\bullet\dagger$} & \textbf{11.05}\phantom{$\bullet\dagger$} &  17.79\phantom{$\bullet\dagger$} \\
     &  SL POS &  7.75\phantom{$\bullet\dagger$}  &   12.33\phantom{$\bullet\dagger$}  & 7.18\phantom{$\bullet\dagger$} & 12.62\phantom{$\bullet\dagger$}  &  \textbf{13.06}\phantom{$\bullet\dagger$}  & 18.05\phantom{$\bullet\dagger$} &  \textbf{11.00}\phantom{$\bullet\dagger$} & \textbf{17.99}\phantom{$\bullet\dagger$}  \\
     &  SL MSD &  \textbf{8.15}$\bullet\dagger$ &   \textbf{12.55}\phantom{$\bullet\dagger$} &  \textbf{7.35}$\bullet$\phantom{$\dagger$} &   \textbf{13.09}$\bullet\dagger$  &  \textbf{13.39}$\bullet$\phantom{$\dagger$} & 18.08\phantom{$\bullet\dagger$} & \textbf{11.70}$\bullet\dagger$  &  \textbf{18.71}$\bullet\dagger$  \\
     &  TL DUM & 7.64\phantom{$\bullet\dagger$} &  12.14\phantom{$\bullet\dagger$}  &   6.49\phantom{$\bullet\dagger$} & \textbf{12.75}\phantom{$\bullet\dagger$} & 12.69\phantom{$\bullet\dagger$} & 17.94\phantom{$\bullet\dagger$}  & 10.78\phantom{$\bullet\dagger$}  & \textbf{18.62}\phantom{$\bullet\dagger$}  \\
     &  TL POS &   \textbf{8.11}$\bullet$\phantom{$\dagger$} &  \textbf{12.58}$\bullet$\phantom{$\dagger$}  & \textbf{7.50}$\bullet\dagger$ &  \textbf{13.68}$\bullet\dagger$  &    \textbf{13.38}$\bullet$\phantom{$\dagger$} &   \textbf{18.76}$\bullet$\phantom{$\dagger$} &  \textbf{11.14}\phantom{$\bullet\dagger$} & \textbf{18.33}\phantom{$\bullet\dagger$}  \\
    &  TL MSD &  \textbf{8.08}$\bullet$\phantom{$\dagger$} &   \textbf{12.41}\phantom{$\bullet\dagger$} & 7.14$\bullet$\phantom{$\dagger$} &\textbf{12.98}\phantom{$\bullet\dagger$}  &    \textbf{13.13}$\bullet$\phantom{$\dagger$} &  \textbf{18.54}$\bullet$\phantom{$\dagger$} &  \textbf{11.23}$\bullet$\phantom{$\dagger$} &  \textbf{18.45}\phantom{$\bullet\dagger$} \\
    &  SL MSD/TL POS &  \textbf{9.02}\phantom{$\bullet\dagger$} &  \textbf{13.51}\phantom{$\bullet\dagger$} & \textbf{7.95}\phantom{$\bullet\dagger$}   &   \textbf{13.27}\phantom{$\bullet\dagger$} &  \textbf{14.61}\phantom{$\bullet\dagger$} &   \textbf{19.40}\phantom{$\bullet\dagger$}  & \textbf{11.65}\phantom{$\bullet\dagger$}   &  \textbf{19.35}\phantom{$\bullet\dagger$} \\
    \hline
    \multirow{10}{*}{German} &  None (baseline)&  17.17\phantom{$\bullet\dagger$} & 26.79\phantom{$\bullet\dagger$}  & 15.07\phantom{$\bullet\dagger$}   & 28.32\phantom{$\bullet\dagger$} & 18.52\phantom{$\bullet\dagger$}  & 27.61\phantom{$\bullet\dagger$}  & 14.97\phantom{$\bullet\dagger$} & 27.73\phantom{$\bullet\dagger$}  \\
    &  SL DUM & 17.19\phantom{$\bullet\dagger$}  &   \textbf{27.40}\phantom{$\bullet\dagger$} & 14.98\phantom{$\bullet\dagger$} &  27.97\phantom{$\bullet\dagger$} & 18.85\phantom{$\bullet\dagger$} & 27.98\phantom{$\bullet\dagger$}  & 15.35\phantom{$\bullet\dagger$} & 27.90\phantom{$\bullet\dagger$} \\
    &  SL POS &  \textbf{18.43}$\bullet$\phantom{$\dagger$}  &  \textbf{27.78}\phantom{$\bullet\dagger$} & \textbf{16.34}$\bullet$\phantom{$\dagger$} &  \textbf{29.16}$\bullet$\phantom{$\dagger$}  &   \textbf{19.63}$\bullet$\phantom{$\dagger$} &  27.99\phantom{$\bullet\dagger$} &  \textbf{15.68}\phantom{$\bullet\dagger$}  &  \textbf{28.32}$\bullet$\phantom{$\dagger$}  \\
    &  SL MSD &   \textbf{18.41}$\bullet$\phantom{$\dagger$} &   \textbf{27.52}\phantom{$\bullet\dagger$} & \textbf{16.36}$\bullet$\phantom{$\dagger$}  &  \textbf{29.01}$\bullet$\phantom{$\dagger$} &   \textbf{18.41}$\bullet$\phantom{$\dagger$} &  27.97\phantom{$\bullet\dagger$} & \textbf{15.86}$\bullet$\phantom{$\dagger$}  &  \textbf{27.98}\phantom{$\bullet\dagger$} \\
    &  TL DUM & 17.77\phantom{$\bullet\dagger$} &  \textbf{27.50}\phantom{$\bullet\dagger$} & 15.17\phantom{$\bullet\dagger$}  &  27.67\phantom{$\bullet\dagger$}  & 17.77\phantom{$\bullet\dagger$} & 27.96\phantom{$\bullet\dagger$} & 15.17\phantom{$\bullet\dagger$}  &  27.67\phantom{$\bullet\dagger$}  \\
     &  TL POS &  \textbf{17.97}\phantom{$\bullet\dagger$} &   \textbf{28.06}$\bullet\dagger$ &  \textbf{15.48}$\dagger$\phantom{$\bullet$}  &  \textbf{29.07}$\bullet\dagger$  &  \textbf{17.97}\phantom{$\bullet\dagger$} & \textbf{28.28}$\bullet$\phantom{$\dagger$} &  \textbf{15.48}$\bullet\dagger$  &  \textbf{29.07}$\bullet$\phantom{$\dagger$}  \\
   &  TL MSD &  17.53\phantom{$\bullet\dagger$}  & 27.22\phantom{$\bullet\dagger$} & 14.91\phantom{$\bullet\dagger$} &   28.34$\bullet$\phantom{$\dagger$} &  17.53\phantom{$\bullet\dagger$}  & \textbf{28.45}$\bullet$\phantom{$\dagger$} & 14.91\phantom{$\bullet\dagger$} &   28.34\phantom{$\bullet\dagger$} \\
    &  SL MSD/TL POS &  \textbf{20.44}\phantom{$\bullet\dagger$} &  \textbf{29.31}\phantom{$\bullet\dagger$}  & \textbf{17.20}\phantom{$\bullet\dagger$} &  \textbf{29.46}\phantom{$\bullet\dagger$} &  \textbf{20.44}\phantom{$\bullet\dagger$} &  \textbf{29.78}\phantom{$\bullet\dagger$}  & \textbf{17.20}\phantom{$\bullet\dagger$} &  \textbf{29.46}\phantom{$\bullet\dagger$} \\
    \cline{2-10}
         & WMT None (baseline)& - &  37.97\phantom{$\bullet\dagger$}   & - & 38.59\phantom{$\bullet\dagger$}  & - &  39.94\phantom{$\bullet\dagger$}  & - &  40.12\phantom{$\bullet\dagger$} \\
        & WMT SL MSD/TL POS & - & \textbf{39.51}\phantom{$\bullet\dagger$}  & - & 38.48\phantom{$\bullet\dagger$} & - & \textbf{40.61}\phantom{$\bullet\dagger$}   & - & 40.40\phantom{$\bullet\dagger$}  \\
    \hline
    \multirow{8}{*}{Spanish} &  None (baseline)& 20.75\phantom{$\bullet\dagger$} &  26.75\phantom{$\bullet\dagger$} & 19.18\phantom{$\bullet\dagger$}   &  26.78\phantom{$\bullet\dagger$} & 19.88\phantom{$\bullet\dagger$} & 25.59\phantom{$\bullet\dagger$} &  18.03\phantom{$\bullet\dagger$} & 25.82\phantom{$\bullet\dagger$}  \\
    &  SL DUM &  \textbf{21.13}\phantom{$\bullet\dagger$}  &  27.03\phantom{$\bullet\dagger$} & 19.28\phantom{$\bullet\dagger$} & 26.63\phantom{$\bullet\dagger$}  & \textbf{20.30}\phantom{$\bullet\dagger$} & \textbf{26.18}\phantom{$\bullet\dagger$}  & 18.46\phantom{$\bullet\dagger$} & 25.80\phantom{$\bullet\dagger$} \\
    &  SL POS &  \textbf{21.88}$\bullet$\phantom{$\dagger$} & 26.92\phantom{$\bullet\dagger$}  &  \textbf{19.57}\phantom{$\bullet\dagger$}  &  27.11$\bullet\dagger$  & \textbf{20.78}$\bullet$\phantom{$\dagger$} &  \textbf{26.17}\phantom{$\bullet\dagger$} & \textbf{18.72}\phantom{$\bullet\dagger$}  & 25.77\phantom{$\bullet\dagger$}  \\
    &  SL MSD &  \textbf{21.68}$\bullet$\phantom{$\dagger$} &   \textbf{27.29}$\dagger$\phantom{$\dagger$} &  \textbf{20.00}$\bullet\dagger$  &  26.60\phantom{$\bullet\dagger$} & \textbf{20.49}\phantom{$\bullet\dagger$} & \textbf{26.22}\phantom{$\bullet\dagger$}  & \textbf{18.55}\phantom{$\bullet\dagger$}  & 25.82\phantom{$\bullet\dagger$}  \\
    &  TL DUM & 20.93\phantom{$\bullet\dagger$} & 26.79\phantom{$\bullet\dagger$}  &  \textbf{19.57}\phantom{$\bullet\dagger$}  &  25.95\phantom{$\bullet\dagger$}  & 19.86\phantom{$\bullet\dagger$} & 25.66\phantom{$\bullet\dagger$}  &  18.06\phantom{$\bullet\dagger$} & 25.72\phantom{$\bullet\dagger$}  \\
    &  TL POS &  \textbf{21.29}$\bullet$\phantom{$\dagger$} &   27.04\phantom{$\bullet\dagger$} &  \textbf{19.60}\phantom{$\bullet\dagger$}  &  27.02$\bullet$\phantom{$\dagger$} & \textbf{20.91}$\bullet\dagger$ & \textbf{26.17}$\bullet$\phantom{$\dagger$}  & \textbf{18.76}$\bullet$\phantom{$\dagger$}  & 26.09$\dagger$\phantom{$\dagger$} \\
   &  TL MSD &  \textbf{21.43}$\bullet$\phantom{$\dagger$} &  26.91\phantom{$\bullet\dagger$} & \textbf{19.72}\phantom{$\bullet\dagger$}  &  \textbf{27.23}$\bullet$\phantom{$\dagger$} & \textbf{20.45}$\bullet$\phantom{$\dagger$}  & \textbf{26.13}$\bullet$\phantom{$\dagger$} & \textbf{18.82}$\bullet$\phantom{$\dagger$}  & 25.60\phantom{$\bullet\dagger$}  \\
    &  SL MSD/TL POS &   \textbf{22.69}\phantom{$\bullet\dagger$} &  \textbf{27.79}\phantom{$\bullet\dagger$} & 20.35\phantom{$\bullet\dagger$}   &   26.89\phantom{$\bullet\dagger$} &  \textbf{21.87}\phantom{$\bullet\dagger$} &  \textbf{27.02}\phantom{$\bullet\dagger$} &  \textbf{19.07}\phantom{$\bullet\dagger$}  &  \textbf{26.30}\phantom{$\bullet\dagger$} \\
    \hline
    \multirow{8}{*}{Turkish} &  None (baseline)&6.21\phantom{$\bullet\dagger$}  & 9.78\phantom{$\bullet\dagger$}  &  4.24\phantom{$\bullet\dagger$}  & 10.30\phantom{$\bullet\dagger$} & 10.66\phantom{$\bullet\dagger$} & 15.00\phantom{$\bullet\dagger$} & 8.30\phantom{$\bullet\dagger$}  &   15.50\phantom{$\bullet\dagger$}  \\
    &  SL DUM & 6.14\phantom{$\bullet\dagger$}  &  9.78\phantom{$\bullet\dagger$} & 4.31\phantom{$\bullet\dagger$} &  10.60\phantom{$\bullet\dagger$}  & 10.18\phantom{$\bullet\dagger$}  &  15.28\phantom{$\bullet\dagger$} & 8.24\phantom{$\bullet\dagger$}  & 15.90\phantom{$\bullet\dagger$} \\
    &  SL POS &  \textbf{6.98}$\bullet$\phantom{$\dagger$} &  \textbf{10.37}$\bullet$\phantom{$\dagger$}  & 4.38\phantom{$\bullet\dagger$} &   \textbf{10.69}\phantom{$\bullet\dagger$} & 10.65$\bullet$\phantom{$\dagger$} &  15.36\phantom{$\bullet\dagger$} & 8.06\phantom{$\bullet\dagger$}  & \textbf{16.10}\phantom{$\bullet\dagger$}  \\
    &  SL MSD &  \textbf{7.02}$\bullet$\phantom{$\dagger$} &   \textbf{10.80}$\bullet\dagger$ &  \textbf{4.62}$\bullet$\phantom{$\dagger$}  & \textbf{10.69}\phantom{$\bullet\dagger$}  & 10.73$\bullet$\phantom{$\dagger$} & \textbf{15.39}\phantom{$\bullet\dagger$}  &  7.90$\bullet$\phantom{$\dagger$} &  \textbf{15.93}\phantom{$\bullet\dagger$} \\
  &  TL DUM & 5.90\phantom{$\bullet\dagger$}  & 9.96\phantom{$\bullet\dagger$}  & 4.20\phantom{$\bullet\dagger$}  &  10.39\phantom{$\bullet\dagger$} & 10.37\phantom{$\bullet\dagger$} & 14.79\phantom{$\bullet\dagger$}  & 8.32\phantom{$\bullet\dagger$}  & \textbf{16.21}\phantom{$\bullet\dagger$} \\
  &  TL POS &  6.50$\bullet\dagger$ & 10.08$\dagger$\phantom{$\dagger$}  &  4.50$\bullet$\phantom{$\dagger$}  &  \textbf{10.66}\phantom{$\bullet\dagger$} & 10.84$\bullet\dagger$ & 14.56\phantom{$\bullet\dagger$}  &  \textbf{9.43}$\bullet\dagger$ &   \textbf{16.12}\phantom{$\bullet\dagger$}  \\
  &  TL MSD & 5.96\phantom{$\bullet\dagger$} & 9.63\phantom{$\bullet\dagger$} & \textbf{4.59}$\bullet$\phantom{$\dagger$}  &  10.48\phantom{$\bullet\dagger$}  & 10.23\phantom{$\bullet\dagger$} & \textbf{15.90}$\bullet\dagger$  & \textbf{8.75}$\bullet$\phantom{$\dagger$}  &    \textbf{16.42}\phantom{$\bullet\dagger$}  \\
 &  SL MSD/TL POS &  \textbf{7.27}\phantom{$\bullet\dagger$} &  \textbf{10.85}\phantom{$\bullet\dagger$}  & \textbf{4.95}\phantom{$\bullet\dagger$}   &  \textbf{10.84}\phantom{$\bullet\dagger$} & \textbf{12.08}\phantom{$\bullet\dagger$} & \textbf{16.75}\phantom{$\bullet\dagger$}  &  \textbf{9.38}\phantom{$\bullet\dagger$}  & \textbf{16.99}\phantom{$\bullet\dagger$} \\
    \hline
    \end{tabular}
    \caption{
    BLEU scores computed on the test set for the different language pairs, corpus sizes and setups evaluated. See Section \ref{sec:results} (first paragraph) for a description of the different symbols annotating the scores.
    }
    \label{tab:results-bleu-chrF2-slen}
\end{table*}

Table~\ref{tab:results-bleu-chrF2-slen} shows the BLEU~\cite{papineni2002bleu} 
scores obtained by the different systems. 
A score in bold means that the 
system outperforms the baseline (labelled as \emph{None}) by a statistically significant margin. A bullet ($\bullet$) next to the score of a system with interleaved POS or MSD tags means that it outperforms the system with DUM tags in the same language side (SL or TL) by a statistically significant margin.\footnote{\revisarcr{Those systems trained with both SL MSD and TL POS tags
could not be compared with systems with both SL DUM and TL DUM tags because the latter were not included in the experimental set-up in order to keep the experiments to a manageable size. Hence, their scores do not contain any bullet. }} A dagger ($\dagger$) next to the score of a system with POS or MSD tags means that it
outperforms the system with the opposite tag (either MSD or POS) in the same language side by a statistically significant margin. Statistical significance was assessed with 
paired bootstrap resampling~\cite{koehn2004statistical} ($p = 0.05$; $1\,000$ iterations).

As the four languages paired with English are morphologically richer than English, we split the analysis of the results \revisar{we describe next} into two groups: translation into a TL morphologically richer than the SL (pairs with English as SL), and translation from a morphologically richer SL (pairs with English as TL). \revisarcr{It is also worth mentioning that, in all the scenarios evaluated, when a system was trained with interleaved TL tags, the decoder alternately produced TL tags and surface forms at test time as expected. }

\paragraph{Translation into a morphologically rich language.}

When the TL is morphologically richer than the SL, interleaved tags lead to higher BLEU scores, although the impact changes depending on the information encoded in the tag and the language  where they are used (SL or TL).
SL DUM tags are not very effective: they bring a statistically significant increase in BLEU only to 2 out of the 8 systems
evaluated with the recurrent architecture,\footnote{\revisarcr{Four language pairs and two training corpus sizes.}} and to none of the 8 Transformer systems. 
SL POS and MSD tags generally outperform DUM tags, as they contain information that helps to obtain a better representation of the SL sentence and break the grammatical ambiguity of English~\cite{sennrich-haddow-2016-linguistic}. There is a statistically significant difference between SL POS and SL MSD tags in 6 out of the 16 systems evaluated,\footnote{\revisarcr{Four language pairs, two training corpus sizes and two architectures.}} and in 5 out of these 6 systems MSD tags outperform POS tags. For some language pairs and training corpus sizes, enriching the SL representation with information about number, verbal mood, etc. proves to be useful.

We can find stronger differences between the different types of tags 
in the TL. TL DUM tags 
are useful for the recurrent English--German systems, in line with the findings by~\newcite{wagner2017}, but 
their contribution to other language pairs and the Transformer architecture is less clear. 
The most relevant trend is that using only POS tags in the TL consistently outperforms the use of MSD tags: 
statistically significant differences are found in all TLs but Spanish, a Romance language which  
has the simplest morphology. 
This result is further investigated in sections~\ref{sec:analysis} and \ref{sec:tlmorph}. 
Finally, combining SL MSD and TL POS tags 
leads to the highest scores.

\paragraph{Translation from a morphologically rich language.}
The effects of using SL tags when the SL is morphologically richer than the TL are similar to those observed in the opposite scenario:
POS and MSD tags often outperform DUM tags. 
When statistically significant differences between POS and MSD tags are found, they favour MSD tags.
Concerning TL tags, the systematic degradation 
observed for MSD tags is less frequent than in the opposite direction, and it is mainly concentrated in the smallest corpus size. 
An explanation could be that morphological information in English is less complex and easier to predict from the SL sentence. 
Finally, combining SL MSD tags and TL POS tags also leads to the highest scores.

\paragraph{Large-scale training data.}
The results for the English--German WMT large-scale training data, 
also depicted in Table~\ref{tab:results-bleu-chrF2-slen}, 
show a different picture. We can still observe 
that the use of interleaved tags brings a statistically significant improvement, 
but this only happens in the recurrent architecture. Transformer systems do not benefit from the interleaved linguistic annotations when the training corpus size is large.\footnote{\revisar{In order to confirm that the differences observed were not caused by the random initialisation of the network weights, we trained the systems on WMT data two more times. The BLEU scores obtained for the additional training runs confirm that linguistic annotations only help in recurrent systems. 
We did not perform additional training runs for the under-resourced scenario because of time constraints and because the conclusions drawn for that scenario are based on the results obtained for different language pairs (and hence training runs), which we think makes them solid enough.} }
\revisarcr{A potential explanation is provided in the next section.}


\paragraph{Main findings.}
In line with previous works~\cite{nadejde:2017,wagner2017}, the results analysed so far suggest  that 
\revisar{interleaved linguistic annotations}
are helpful both in the SL and the TL and they should be included in both languages in order to maximize performance. While morphological \revisar{features} can be useful in the SL, they should be avoided in the TL if it is morphologically rich. Even when large corpora are available, linguistic \revisar{annotations} can help to boost translation quality.





\section{Error analysis}
\label{sec:analysis}
\begin{figure}[tb]
    \centering
     \includegraphics[width=1.0\textwidth]{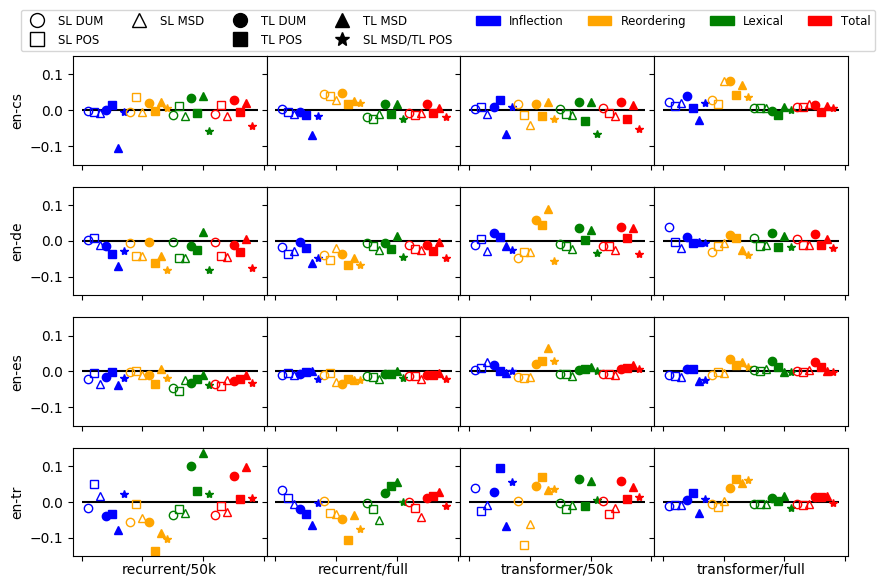}
     \caption{For language pairs with English as SL, relative changes in the number of errors for each error category, training corpus size and type of interleaved tag. }
     \label{fig:hjerson:ensl}
   \end{figure}

\begin{figure}[tb]
    \centering
     \includegraphics[width=1.0\textwidth]{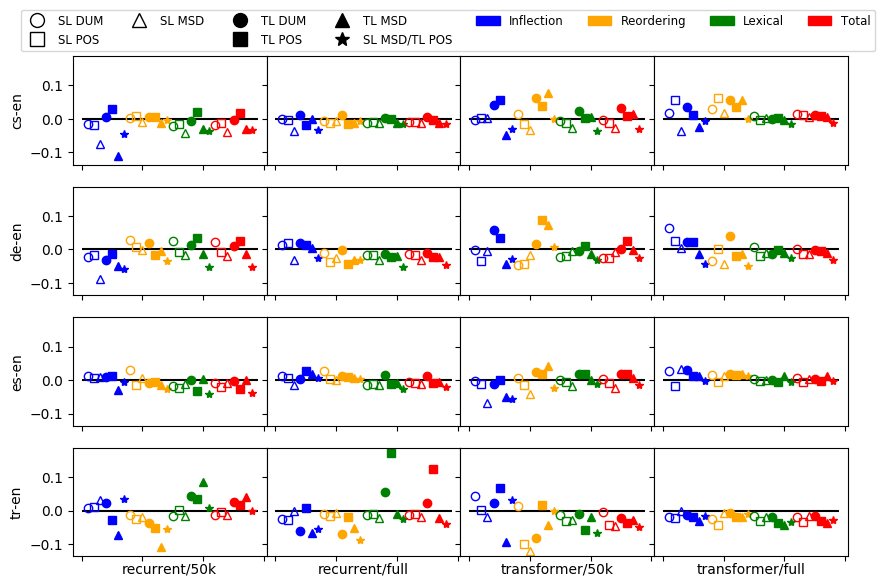}
     \caption{For language pairs with English as TL, relative changes in the number of errors for each error category, training corpus size and type of interleaved tag.}
     \label{fig:hjerson:entl}
   \end{figure}

To better understand the results obtained, we computed the relative difference in the number of Hjerson errors between the systems with interleaved tags and the baseline;
\footnote{Computed as $(\mathrm{\# errors\_interleaved} - \mathrm{\# errors\_baseline})/\mathrm{\# errors\_baseline}$.} 
a positive value means that the system made more errors than the baseline.
As we did before, we split the results into two groups of language pairs: those with English as SL, depicted in Figure~\ref{fig:hjerson:ensl}, and those with English as TL, depicted in Figure~\ref{fig:hjerson:entl}.
In the remainder of this section, we analyse the results obtained and illustrate them with examples. 

%

  \begin{table*}[tb]
\begin{center}
\footnotesize
\begin{tabular}{c|l|l}
\textbf{Pair} & \textbf{Tags} & \textbf{Sentence}  \\
\hline
\multirow{4}{*}{\texttt{en-es}} & orig/ref. & This does not make \textbf{matters simple} for the Germans. $\rightarrow$ Esto no resulta fácil para los alemanes. \\
  &None & Esto no hace \textbf{que los alemanes sean simples}.  \\
 &SL MSD & Esto no hace \textbf{que las cosas sean simples} para los alemanes.  \\
\hline
\multirow{4}{*}{\texttt{en-de}}& orig/ref, & Miller was not engaging in literary criticism  $\rightarrow$ Miller engagierte sich nicht in \textbf{literarischer} Kritik    \\
& TL POS & Miller setzte sich nicht an \textbf{literarische} Kritik  \\
& TL MSD & Miller befasste sich nicht mit \textbf{literarischer} Kritik \\
\hline
\multirow{4}{*}{\texttt{en-de}}& orig/ref &  "The children should help prepare the food." $\rightarrow$ "die Kinder sollen \textbf{helfen}, das Essen zuzubereiten". \\
& TL POS & "die Kinder sollten dabei \textbf{helfen}, die Nahrung vorzubereiten". \\
& TL MSD & "die Kinder sollten dazu \textbf{beitragen}, die Lebensmittel vorzubereiten" \\
\hline
\end{tabular}
\end{center}

\caption{Examples of how interleaving SL or TL tags change the translations. Translations obtained with the recurrent systems trained on the full News Commentary corpus.}
\label{tab:examples}
\end{table*}



\paragraph{SL tags.}
SL POS and SL MSD tags systematically reduce lexical 
errors (green, empty squares and triangles are below the horizontal line). 
Reordering errors are also reduced with the exception of English--Czech, in which the TL has a relatively flexible word order.\footnote{\revisarcr{Improvements in reordering quality are not captured by the automatic error classification if translation hypotheses with an acceptable word order that differs from that in the reference are produced by the system. }} Concerning inflection errors, there is not clear trend.
As aforementioned, a possible explanation 
could be that SL tags help to obtain more accurate representations of the SL sentences; 
since inflection errors are related to modeling TL grammar rather than to representing the SL sentence, they are not reduced by interleaving SL tags.
All these results are compatible with the 
evaluation metrics, which showed that SL tags generally improve translation quality. 
In the first example in Table~\ref{tab:examples}, SL tags help to obtain a better representation of the SL sentence: the system is able to interpret that \emph{matters} is acting as a noun and produces \emph{hace que las cosas sean simples} (en: \emph{it makes things simple}) instead of \emph{hace que los alemanes sean simples} (en: \emph{it makes Germans simple}).


\paragraph{TL tags.}
Different error distributions can be observed depending on the information encoded in the TL tags. 
TL MSD tags systematically reduce inflection errors in both architectures (the blue, filled triangle is usually among the lowest points in the figure). 
The largest inflection error reductions occur with highly inflected TLs such as Czech and Turkish. 
TL POS tags, on the contrary, do not systematically reduce inflection errors. 
Hence, the system using TL MSD tags is  using the morphological features they encode (tense, number, etc.) for producing the correct inflected form \revisarcr{according to the reference}. 
In the second example in Table~\ref{tab:examples}, the  system using TL MSD tags generates the right inflected form of the German word \emph{literarische} because it has predicted its dative case first.

However, the prediction of MSD tags with complex morphology (see Figure~\ref{fig:hjerson:ensl}) also leads to an increase in lexical errors in comparison with the prediction of POS tags. It can be observed that TL MSD tags bring an increase in lexical errors over the baseline (note the green, filled triangles at the top of the figures),
while the impact in lexical errors of the POS tags is less clear. 
Similarly to inflection errors, the difference between the increases of lexical errors brought by MSD and POS tags is larger for Turkish and Czech, which are the two languages with the most sparse MSD tags. Turkish is a Turkic agglutinative language and Czech is a Slavic fusional language with seven cases and four genders.
This is compatible with the automatic evaluation metrics: although using TL MSD tags leads to a more grammatical output, the increase in lexical errors makes the system produce  translations that are \revisarcr{overall} less similar to the reference. Note that lexical errors are the most frequent ones.\footnote{For the baseline recurrent system, the average 
 and standard deviation over the four language pairs with English as SL of the absolute number of errors of each type in the test set, \revisar{expressed in thousands}, are as as follows. Inflection:  4.1 {$\pm$} 0.7, Reordering: 5.3 {$\pm$} 0.6 , Lexical: 28.7 {$\pm$} 2.1. } 
%
%
The third example in Table~\ref{tab:examples} shows that the system with TL MSD tags translates the verb \emph{help} as \emph{beitragen} rather than \emph{helfen}, which is a more precise translation in that context. 

When SL MSD tags and TL POS tags are both interleaved, there is a general  reduction in the three error categories as compared with the systems using only tags in one of the languages. This confirms that the advantages of SL and TL tags are complementary.

%

\paragraph{Differences between architectures.}
Finally, there is a noticeable difference in how the type of errors made by the systems change when interleaving TL tags in recurrent and Transformer architectures.
Reordering errors consistently increase in Transformer systems, while they tend to decrease in recurrent systems.
Moreover, TL DUM tags consistently increase the total number of translation errors when they are added to a Transformer system and the TL is highly inflected (observe the red, filled circle usually above the horizontal line in Figure~\ref{fig:hjerson:ensl}), while their impact is not clear in recurrent systems. These two findings suggest that adding extra tokens to the TL stream is not the best way of introducing linguistic annotations in self-attention-based NMT systems. It could also explain the results for the large-scale WMT data, where only recurrent systems were able to take advantage of the linguistic annotations. \revisar{This hypothesis is also compatible with the results reported on WMT data by \newcite{yang-etal-2019-latent}, who successfully leveraged TL linguistic annotations in Transformer systems using an \emph{ad-hoc} architecture.}

\section{Analysing the effect of target language morphology}
\label{sec:tlmorph}


\begin{figure}[t]
     \centering
        \includegraphics[width=0.49\textwidth]{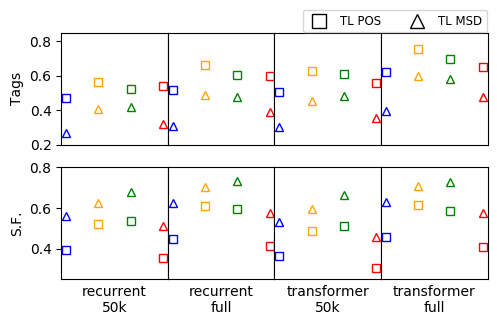}
        \includegraphics[width=0.49\textwidth]{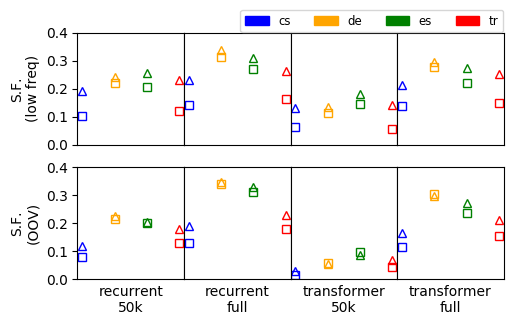}
    \caption{\revisar{For language pairs with English as SL, tag prediction accuracy (labelled as \emph{Tags}) and surface form prediction accuracy (labelled as \emph{S.F.}) forcing, respectively, surface forms and tags from the reference.}}
    \label{fig:force-initial}
\end{figure}

We compare the output of the systems interleaving TL POS and TL MSD tags in order to ascertain whether the increase in lexical errors is caused by the difficulty of predicting the more complex and sparse MSD tags, or by the conditioning of the prediction of  surface forms on TL MSD tags. In the latter case, there is the risk that the system learns to strongly condition on tags and avoids generating new words~\cite{tamchyna-etal-2017-modeling}. This problem could be exacerbated by the sparsity of TL MSD tags because some of them may co-occur only with a few surface forms in the training corpus.

We tried to answer this question by 
independently evaluating 
the prediction of 
tags and surface forms, and
comparing the systems interleaving POS and MSD tags. 
The prediction of surface forms was evaluated by re-decoding the test set and forcing the system to choose the tags from the reference during beam search, whereas the prediction of tags was evaluated by forcing the surface forms from the reference.   
If the system really learned to strongly condition on tags, when a tag was observed together with only a few surface forms in the training corpus, low-frequency words would not be generated when translating the test set. To test this hypothesis 
we studied the surface form prediction accuracy for two subsets: infrequent words (frequency $\le$ 10 in the training set) 
and out-of-vocabulary (OOV) words. 
Figure~\ref{fig:force-initial} shows the results for those language pairs with English as SL. It can be observed that there is a trade-off between tag and surface form prediction accuracy: MSD tags are more difficult to predict, but conditioning on them leads to better surface form prediction. 
On low-frequency and OOV words, MSD tags still outperform POS tags in terms of surface form prediction accuracy, although the difference between them is smaller.

\begin{figure}
    \centering
       \subfloat[English as SL.\label{subfig-2:dummy}]{%
        \includegraphics[width=0.5\textwidth]{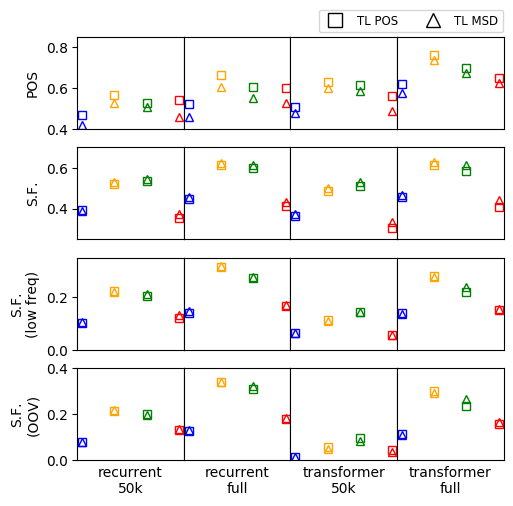}
     }
        \subfloat[English as TL.\label{subfig-3:dummy}]{%
        \includegraphics[width=0.5\textwidth]{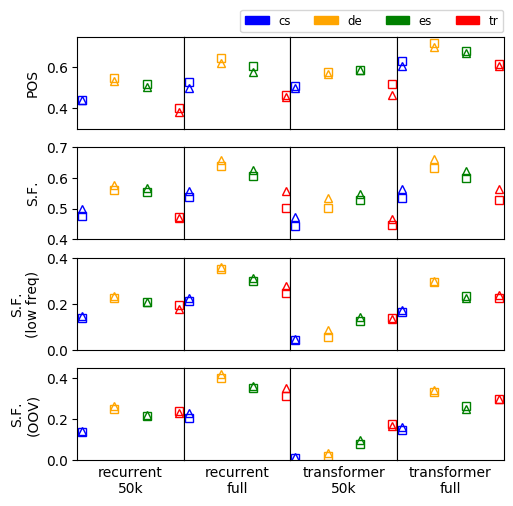}
     }
    \caption{\revisar{
    POS prediction accuracy (labelled as \emph{POS}) and surface form prediction accuracy (labelled as \emph{S.F.}) forcing, respectively, surface forms and POS tags from the reference.}}
    \label{fig:force}
\end{figure}


For a fair comparison of both types of tags, we computed the part-of-speech prediction accuracy when predicting MSD tags.
The results are depicted in the rows labelled as \emph{POS} in Figure~\ref{fig:force}. For highly inflected TLs, those systems that predict MSD tags have consistently lower POS prediction accuracy than those that predict only POS tags. The difference is larger for recurrent systems. These results suggest that the difficulty of predicting together the part of speech and its morphological features is indeed one of the reasons behind the lexical degradation of systems using MSD tags. Sparseness of MSD tags seems to play an important role in this degradation: highly inflected TLs present the largest degradation.

To evaluate the impact of predicting morphological features regardless of the low part-of-speech accuracy of MSD tags, we re-computed surface form prediction accuracy by letting the beam search algorithm choose among those MSD tags with the part of speech in the reference. 
The results, depicted in the rows labelled as \emph{S.F.} in Figure~\ref{fig:force},
show that, if the systems with interleaved MSD tags correctly predicted the part of speech,
the surface form predictions would not be worse than those of systems with interleaved POS tags, 
neither in general nor for low-frequency and out-of-vocabulary words.
Hence, errors in surface form prediction arising from 
strongly conditioning on sparse MSD tags
do not seem to be the main cause behind the degradation of translation quality 
\revisar{brought by TL MSD tags}.
Actually, when tags with the part of speech of the reference are chosen, conditioning on MSD tags outperforms conditioning on POS tags in terms of overall surface form prediction accuracy for  language pairs with English as TL. For the other language pairs, the gain introduced by MSD tags is less clear. One possible reason could be that BPE segmentation does not allow the system to learn a general mapping between tags and word endings from the training data. 
Another explanation could be related to the fact that predicting the morphological gender for German, Czech and Spanish forces the tag prediction task to be aware of TL lexical information, preventing an optimum division of labour between tag and surface form predictions.

In conclusion, the prediction of TL morphological information needs to be factorised 
differently in order not to harm part-of-speech prediction.
For instance, in the morphological analysis field, \newcite{chaudhary-etal-2019-cmu} and \newcite{straka-etal-2019-udpipe} predict part of speech and each morphological attribute independently.

\section{Concluding remarks}

In this paper, we have studied the effects of using linguistic annotations of SL and TL words in under-resourced NMT by interleaving linguistic tags for different language pairs, architectures, training data sizes and types of linguistic information (part of speech and morpho-syntactic descriptions).

We have shown that both SL and TL linguistic annotations are useful, in line with previous works in the literature~\cite{wagner2017}.
SL linguistic annotations lead to more accurate SL sentence representations, and for some language pairs, the use of morpho-syntactic descriptions (consisting of part of speech and morphological features) improves the representation obtained when only part-of-speech tags are used. Surprisingly, for highly inflected TLs, TL linguistic annotations are more useful if they simply consist of part-of-speech information. Using  morpho-syntactic descriptions leads to an overall translation quality degradation in terms of automatic evaluation metrics, even though it improves the grammaticality of the output. We have also shown that  predicting TL morpho-syntactic descriptions frequently results in wrong part-of-speech predictions. Hence, to optimize the use of TL morphological information in NMT, it is advisable to avoid the prediction of part-of-speech and morphological features together as monolithic tags.


The gain introduced by linguistic information encoded as interleaved tags scales to large data availability scenarios only for the recurrent architecture. This result, together with the conclusions of the automatic error analysis, suggest that adding extra tokens to the TL stream is not the optimum way of introducing additional linguistic information in self-attention-based NMT systems.
 
In summary, the use of morpho-syntactic descriptions in the SL and part-of-speech information in the TL, which can be easily obtained even in under-resourced scenarios, systematically improves translation quality when they are simply interleaved in training data as linguistic tags, even without using a morphological generator~\cite{tamchyna-etal-2017-modeling}, which could be error-prone for under-resourced languages, and without any kind of information about syntactic structures.

\section*{Acknowledgements}
Work funded by the European Union’s Horizon 2020 research and innovation programme under grant agreement number 825299, project Global Under-Resourced Media Translation (GoURMET). We thank Mikel L. Forcada for his help characterising the languages used in this paper.

\bibliographystyle{coling}
\bibliography{coling2020}

\appendix
\newpage
\section{Training details}
\label{se:hyperparam}

The optimum training hyper-parameters were obtained by following the grid search process depicted next. At each step, we chose the hyper-parameters that maximised BLEU on the development set. Table~\ref{tab:hyper:rec} shows the optimum hyper-parameters for each language pair and training corpus size for the recurrent architecture while Table~\ref{tab:hyper:trans} shows the same information for the Transformer architecture.

\begin{itemize}
\item First, we explored the optimum number of BPE operations among the following values: 5,000, 10,000, 20,000, and 40,000. 
The rest of hyper-parameters were set to the values recommended by \newcite{sennrich-EtAl:2017:WMT} for the recurrent architecture and by \newcite{NIPS2017_7181} for the Transformer architecture (``base'' configuration), respectively.
\item With the optimum number of BPE operations, we then tested if better results could be obtained with tied embeddings~\cite{press-wolf-2017-using} in the decoder.
\item Finally, we explored the following combinations of hyper-parameter values for each architecture with the best number of BPE operations and tied embedding configuration obtained in the previous steps.
\begin{description}
\item[recurrent:] Hidden and embedding sizes:$(1024,512)$, $(512,512)$, $(512,256)$ and $(256,256)$. 
\item[Transformer:] Model size, number of layers in the encoder and the decoder and number of attention heads: 
$(512,6,8)$, $(256,4,4)$, $(128,2,2)$. 
\end{description}
\end{itemize}

\begin{table*}[tb]
\centering
\footnotesize \addtolength{\tabcolsep}{-1pt}
    \begin{tabular}{c||r|r||r|r}
    \multirow{2}{*}{\textbf{Language}} &  \multicolumn{2}{c||}{\bf English as SL} & \multicolumn{2}{c}{\bf English as TL}  \\
    \cline{2-5}
    & \multicolumn{1}{c|}{\textbf{50k}}  &  \multicolumn{1}{c}{\textbf{Full}}   &  \multicolumn{1}{c|}{\textbf{50k}}  &  \multicolumn{1}{c}{\textbf{Full}} \\
    \hline
     Czech &  $5\,000$;False;$512;512$ &  $10\,000$;True;$1024;512$ &  $5\,000$;False;$1024;512$  & $10\,000$;False;$1024;512$  \\
     German &  $5\,000$;True;$1024;512$ &  $10\,000$;True;$1024;512$ &  $5\,000$;False;$1024;512$  & $10\,000$;False;$1024;512$ \\
     Spanish &  $5\,000$;False;$1024;512$ &  $5\,000$;True;$1024;512$ &  $5\,000$;True;$1024;512$  & $5\,000$;False;$1024;512$ \\
     Turkish &  $5\,000$;False;$1024;512$ &  $10\,000$;False;$1024;512$ &  $5\,000$;True;$1024;512$  & $10\,000$;False;$1024;512$ \\
    \hline
 \end{tabular}
    \caption{
    Optimum hyper-parameters for recurrent models. Each cell contains, in this order: number of BPE operations, whether TL input and output embeddings are shared, size of the hidden layer and size of the embedding layer.
    }
    \label{tab:hyper:rec}
\end{table*}

\begin{table*}[tb]
\centering
\footnotesize \addtolength{\tabcolsep}{-1pt}
    \begin{tabular}{c||r|r||r|r}
    \multirow{2}{*}{\textbf{Language}} &  \multicolumn{2}{c||}{\bf English as SL} & \multicolumn{2}{c}{\bf English as TL}  \\
    \cline{2-5}
    & \multicolumn{1}{c|}{\textbf{50k}}  &  \multicolumn{1}{c}{\textbf{Full}}   &  \multicolumn{1}{c|}{\textbf{50k}}  &  \multicolumn{1}{c}{\textbf{Full}} \\
    \hline
     Czech &  $20\,000$;True;$256;4;4$ & $10\,000$;True;$256;4;4$ &  $20\,000$;True;$256;4;4$  & $20\,000$;True;$256;4;4$ \\
     German &  $5\,000$;True;$256;4;4$ & $10\,000$;True;$256;4;4$ &  $20\,000$;True;$256;4;4$ & $10\,000$;True;$256;4;4$ \\
     Spanish &  $5\,000$;True;$256;4;4$ & $10\,000$;True;$512;6;8$ &  $5\,000$;True;$256;4;4$ & $10\,000$;True;$512;6;8$ \\
     Turkish &  $20\,000$;True;$128;2;2$ & $10\,000$;True;$256;4;4$ &  $5\,000$;True;$128;2;2$ & $20\,000$;True;$256;4;4$ \\
    \hline
 \end{tabular}
    \caption{
    Optimum hyper-parameters for Transformer models.  Each cell contains, in this order: number of BPE operations, whether TL input and output embeddings are shared, size of the model, number of layers and number of attention heads.
    }
    \label{tab:hyper:trans}
\end{table*}

For all systems trained, we applied label smoothing with a value of 0.1 and dropout of 0.1. Unlike~\newcite{sennrich-zhang-2019-revisiting}, we did not use a lexical model neither word dropout. The optimisation algorithm was Adam \cite{Adam} with the inverse square root learning rate decay \cite[Sec.~5.3]{NIPS2017_7181} and 8,000 warm-up iterations. Learning rates were initialised to 0.0004 for recurrent and to 0.0003 for Transformer. Training stopped after 10 validations without any perplexity improvement on the development corpus; validations were performed every 1,000 mini-batches. The model finally used is the one for which the best BLEU score was obtained on the development corpus.
We used the same amount of sentences per mini-batch for all the models trained for a given TL and corpus size; the amount of sentences in each mini-batch ensures that, when SL and TL MSD tags are interleaved, the amount of tokens is below 4,500.

\end{document}